\documentclass[11pt,letterpaper]{article}
\usepackage[style=base]{caption}
\usepackage{cogsys}
\usepackage[T1]{fontenc}
\usepackage{times}
\usepackage[pdftex]{graphicx} 
\usepackage{subfig}

\usepackage{natbib}
\cogsysheading{10}{2022}{1-18}{5/2020}{12/2022}

\ShortHeadings{Personalized Computational Models of Learning}
              {C. J.\ MacLellan, K.\ Stowers, and L.\ Brady}

\begin{document} 

\title{Evaluating Alternative Training Interventions Using \\Personalized Computational Models of Learning}
 
\author{Christopher James MacLellan}{cmaclell@gatech.edu}
\address{Teachable AI Lab, Georgia Institute of Technology, 
         Atlanta, GA 30332 USA}
\author{Kimberly Stowers}{kim.stwrs@gmail.com}
\address{Tim Fletcher Company, Windsor, Ontario, CA}
\author{Lisa Brady}{lisa.brady@selu.edu}
\address{Southeastern Louisiana University, 
         Hammond, LA 70402 USA}
\vskip 0.2in
 
\begin{abstract}
Evaluating different training interventions to determine which produce the best learning outcomes is one of the main challenges faced by instructional designers. 
Typically, these designers use A/B experiments to evaluate each intervention; however, it is costly and time consuming to run such studies.
To address this issue, we explore how computational models of learning might support designers in reasoning causally about alternative interventions within a fractions tutor.
We present an approach for automatically tuning models to specific individuals and show that personalized models make better predictions of students' behavior than generic ones.
Next, we conduct simulations to generate counterfactual predictions of performance and learning for two students (high and low performing) in different versions of the fractions tutor. 
Our approach makes predictions that align with previous human findings, as well as testable predictions that might be evaluated with future human experiments.
\end{abstract}

\section{Introduction} 

The primary goal of a training intervention designer is to improve the performance of an individual or team along some desirable dimension. Interventions might target capabilities for a wide range of tasks across the physical, cognitive, or social domains. They might also target performance across time scales, from seconds and minutes to weeks and months. For example, soldiers might undergo specialized fitness training to increase the weight they can lift, K12 students might complete practice problems to improve their ability to correctly add fractions, and astronauts might practice working in different roles to foster their ability to perform well as team.

Regardless of the task, domain, or time scale, identifying the interventions that best achieve the desired performance goals and evaluating their effectiveness in a cost-effective way is a central challenge for intervention designers. \citet{koedinger2013instructional} sketch out the design space for educational interventions and claim that there are over 200 trillion possible combinations, even when considering just a small design space with 15 possible instructional techniques, three dosage levels, and different dosage choices for early and late instruction. How then does a designer evaluate alternative interventions and select from this enormous set of options in an informed way?

Randomized A/B experiments are the gold standard for evaluating the causal impact of different interventions and quantifying their effectiveness over a baseline (``control'') group.
Unfortunately, running controlled studies is a costly endeavour; getting approval for human experiments, organizing the studies, and running them is no small task. There are also many limitations of A/B studies. Experiments often only compare a few interventions (typically two or three) and it is very difficult to generalize from these interventions to other alternatives. The end result is that A/B experimentation reduces intervention design to a game of twenty questions with nature \citep{newell1973you}, where each question is expensive to answer and only provides one bit of information regarding which option is best. Further, they often treat interventions as one-size-fits-all solutions, when in reality different interventions often have different effects for different people. For example, it is well known that novices learn more from studying worked examples than from engaging in problem solving, but this relationship reverses as students gain more expertise \citep{kalyuga2003}. Accounting for individual differences typically requires more experimental conditions and increased cost, but not properly accounting for these differences when intervening can hinder performance gains.

Given the costs and limitations of A/B experiments, we need computational tools to support teachers, personal trainers, managers, researchers, and other intervention designers in cost effectively selecting options from the range of alternatives. To address this need, we propose to use computational models of human learning. Similar to how bridge designers use parametric analysis to computationally simulate and test bridges prior to deploying them in the real world, we propose to use computational models to simulate and test cognitive training interventions prior to running costly human experiments. 

\section{Background}
As a starting point, we should look to the substantial prior work on student modeling within intelligent tutoring systems. 
For example, early research by \citet{conati-student-modeling} explored the use of Bayesian models to infer students' mental plans and knowledge state based on their observable actions.
One key shortcoming of their work was that it did not account for learning (i.e., how a student's knowledge state changes as a result of practice), effectively treating each practice opportunity as an assessment rather than a learning event.
Another approach by Corbett and Anderson (\citeyear{corbett1994knowledge}), Bayesian Knowledge Tracing, overcomes this issue by using a hidden Markov model to track how a student's knowledge changes in response to practice.
Over the past two decades, researchers have built on this early work, exploring different statistical accounts of student learning \citep{maclellan2015accounting}. These approaches, often referred to as {\it knowledge tracing}, have been successfully applied within tutoring systems to estimate students' knowledge and predict their performance.

Despite this success in guiding personalization and adaptation within tutors, they have shortcomings for our current objective.
First, they require preexisting student data for parameter estimation before they can generate predictions about performance.
This limits their ability to generalize to novel training interventions for which no previously collected data are available.
Additionally, these approaches only model learning and performance abstractly. For example, Bayesian knowledge tracing can predict what knowledge a student has and whether they will get a particular item correct, but it lacks a mechanistic account that actually solves the item. 

In contrast, {\it computational models of learning} can actually perform the task.
Prior systems in this category, such as Cascade \citep{vanlehn1992cascade}, STEPS \citep{ur1995steps},  SimStudent \citep{li2012efficient}, and Apprentice Learner \citep{maclellan2017thesis}, have successfully accounted for students' learning and decision making within tutors.
For example, the Apprentice learner system \citep{maclellan2016apprentice,maclellan2017thesis} embeds a cognitive learning theory in a unified computational framework to generate theory-driven predictions about human performance and learning for alternative interventions, even when no human data are available.
This makes it possible to generate predictions for previously tested training interventions, as well as counterfactual interventions for which data do not yet exist.



In addition to providing a means of evaluating counterfactual interventions, this approach can address the one-size-fits-all problem faced by A/B experiments. Agents can be customized to specific individuals and their unique characteristics (e.g., novices or experts) to better predict how they will be affected by different interventions. 
We refer to this process as creating {\it personalized} models.\footnote{Others have referred to this process as creating {\it individualized} models \citep{jones1992fine,zhang2014understanding}, but we prefer the term {\it personalized} given the prior work using student models to enable personalization within tutors \citep[e.g.,][]{conati2013student}.}
To build on this idea, we explore a novel approach to leveraging performance data when they are available (e.g., from a previous experiment evaluating one possible intervention) to automatically construct personalized agents. We explore whether this enables better prediction of the target individual's performance than generic agents. We also explore their use for generating predictions of how individuals would have uniquely responded to counterfactual training interventions.

In this paper, we explore the use of personalized computational models of learning to support training intervention design and provide evidence for three high-level claims:
\vskip 0.05in

\cbullet
Designers can use these systems to predict how different training interventions will causally affect student learning, even when no prior human data are available;

\cbullet 
If performance data are available for a particular student, then designers can use them to create personalized agents that better predict the student's behavior than a generic one; and 

\cbullet 
Designers can use these agents to generate plausible counterfactual predictions for how individuals will respond to different interventions.

\vskip 0.05in
\noindent
To support these claims, we first describe the fraction arithmetic learning environment we used in our simulations. We then outline our computational account of learning on this task. Next, we discuss our approach for automatically creating personalized agents and present evidence that they predict human performance better than generic agents. Finally, we apply our personalization approach to two students (high and low performing) and use the resulting models to predict plausible counterfactual learning curves for each student across three different interventions. Given that these predictions are counterfactual---that is, we are making predictions for interventions that were not evaluated in the human data---no ground truth data exists to evaluate them. However, we assess their qualitative plausibility  and show they have reasonable agreement with previous findings.

\section{Fraction Arithmetic Tutor}

\begin{figure}[t]
\vskip 0.05in
\begin{center}
\includegraphics[width=\textwidth]{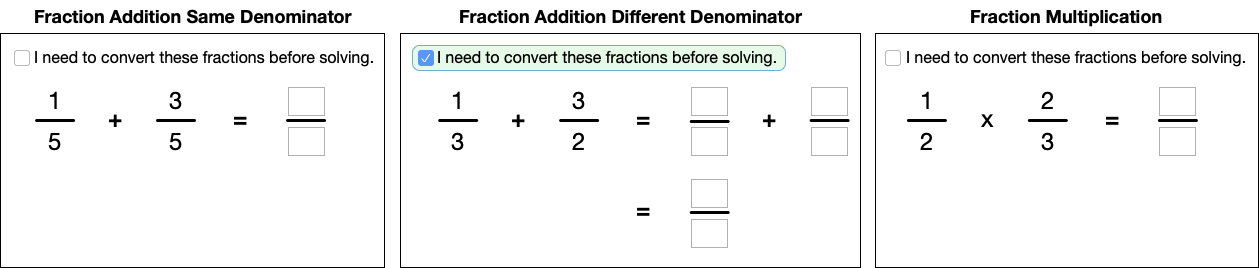}
\caption{Three types of problems presented to students by the fractions tutoring system.} 
\label{fig:fractions-tutor}
\end{center}
\vskip -0.2in
\end{figure} 

To investigate the use of computational models to support design of training interventions, we chose to investigate decision making and learning within a tutoring system for fraction arithmetic. \citet{patel2016block} created this tutor to teach students how to solve three types of problems: fraction addition with same denominators, fraction addition with different denominators, and fraction multiplication. Figure \ref{fig:fractions-tutor} shows the tutoring system interface for each problem type. Following the standard design for intelligent tutoring systems \citep{vanlehn2006behavior}, it provides immediate correctness feedback on each step and students can only proceed once they have performed all steps correctly. Additionally, if students get stuck, then they can request a ``hint'' and the tutor provides an example of how to perform the next step. 

The system scaffolds students in solving these problems in a particular fashion. For all three problem types, they must decide whether to convert the fractions before solving. If they elect to convert and the tutor determines it is appropriate, then it presents them with additional input fields to support conversion, as shown in the middle image of Figure \ref{fig:fractions-tutor}. When adding fractions with same denominator and multiplying them, students can input numerators and denominators in any order and can only mark the problem as done once both fields have correct inputs. When students are solving addition problems with different denominators, they must convert the fraction to common denominators before proceeding. In this case, the tutor requires students to use the butterfly method--the two denominators are multiplied to get a common denominator and the opposing numerators and denominators are multiplied to get new numerators. Additionally, students must input the converted fraction values in a particular order. First they must input the lower left denominator and then they can input the right denominator and the left numerator. Finally, students can enter the right numerator. Once they have converted the fraction, they can input the answers for numerators and denominators in any order. The students can proceed once both answer fields have correct inputs.\footnote{We are not committed to requiring students to use the butterfly strategy or to enter steps in this fixed order, but Patel et al.'s (\citeyear{patel2016block}) original human tutor had these requirements, so we mirrored them in our simulations.} 

For our analysis, we used the ``Study 2'' data from the publicly available ``Fraction Addition and Multiplication'' data set accessed via DataShop \citep{koedinger2010datashop}. These data come from an experiment conducted by \citet{patel2016block} to investigate whether it is better to block or interleave students' fractions practice. For this study, 118 sixth graders were randomly assigned to receive 48 practice problems in either a blocked or interleaved order. Half of the students in the blocked condition received all addition problems with same denominators, then all addition problems with different denominators, then all multiplication problems. The other half received all multiplication problems, then all addition problems with same denominators, then all addition problems with different denominators. Each block presented problems in random order. Students in the interleaved condition received a randomized ordering of all problems. The main finding of this study was that students have lower error during practice in the blocked condition, but better posttest performance in the interleaved condition, suggesting that interleaving practice on fraction arithmetic yields better learning than blocking.

\section{Apprentice Learner Architecture} \label{sec:apprentice}

\begin{figure}[t!]
\vskip 0.05in
\begin{center}
\includegraphics[width=0.7\textwidth]{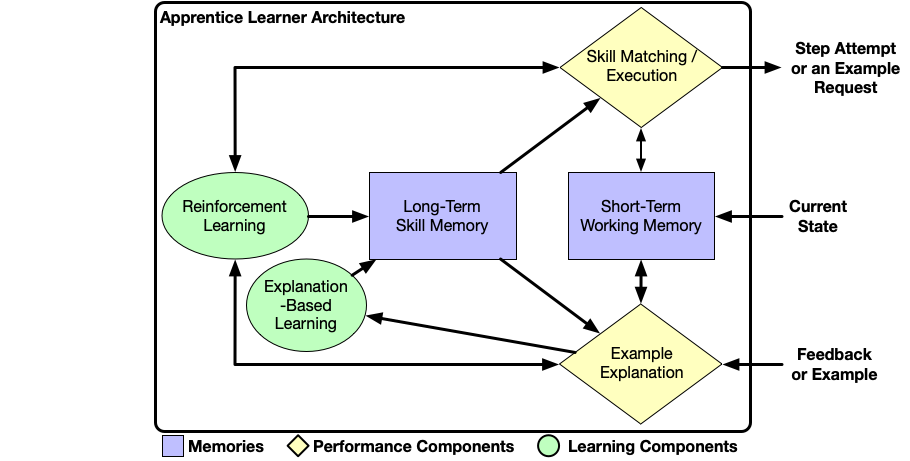}
\caption{A conceptual depiction of the memories (blue boxes), performance components (yellow diamonds), and learning components (green circles) for our Apprentice Learner agents.} 
\label{fig:al-model}
\end{center}
\vskip -0.2in
\end{figure}

To account for human learning in the fractions tutor, we constructed agents using the the Apprentice Learner Architecture \citep{maclellan2016apprentice,maclellan2017thesis}, which provides a framework for modeling human learning and decision making within tutors. Our architecture, which Figure \ref{fig:al-model} depicts conceptually, has two memories: a long-term store that contains skills (hand authored or learned) and a short-term store with working memory elements that are manipulated through skill execution. Skills have three parts: conditions that constrain when the skill applies, effects that update the working memory when it is executed, and a function that predicts the expected future reward of executing the skill in a given state. 

The architecture has two performance components that support interaction with and learning from tutoring systems. First, the skill matching and execution component matches skills from long-term memory against current working memory elements. When multiple skills match, it uses the expected value function associated with each skill to predict the future reward that will be achieved by executing the respective skill in the current state. The skill with the highest expected reward is selected for execution. When an agent fires a skill that produces only internal changes in working memory, it gets a small penalty (e.g., $-0.01$) to discourage unnecessary action. When it executes a skill that generates an external step within the tutoring system, it receives feedback on its action ($+1$ for a correct step and $-1$ for an incorrect step). If an agent has no actions that are predicted to produce reward, then it requests a hint from the tutor.

When the system receives feedback on its actions or a worked example, it invokes an explanation component. When receiving feedback, it already has a trace in its working memory to explain the last action (from generating the step initially). 
However, when explaining an example where no previously generated trace is available, the agent creates one by searching for a chain of skills from long-term memory that explain the example. 
Unlike skill matching and execution, the example explanation process considers all skills, regardless of whether they are predicted to generate reward in the current state, so that it can identify and learn new behaviors.
Once the agent has constructed an explanation, it compiles it into a new skill (akin to macro-operator learning) using a form of explanation-based learning \citep{dejong1986explanation}. This new skill can then be used in subsequent problem solving and learning.
While engaging in skill execution and example explanation, the system uses Q-learning \citep{mnih2015human} to update the expected value function associated with each skill whenever it is fired.

As an example, imagine an agent faced with solving the problem {\it What is 2+3?} In this case, a match is made between learned skills and the current problem state, with the agent executing the one with the highest predicted reward in the current state. If no skills with predicted reward match (a typical initial response), then the agent requests a demonstration from the tutor (e.g., the tutor enters a 5 in the answer field). The agent then leverages its existing skills to search for an explanation of this demonstration, during which it executes skills even if they are predicted to result in penalties. Using this process, the agent might explain the demonstration as two steps (adding the first and second numbers and copying the result into the answer field). Next it compiles this explanation into a new skill (one that performs both adding and copying). 
Throughout this process the agent uses reinforcement learning to update each skill's expected value function whenever the skill is fired, which lets it prioritize matching skills and determine which should be applied in a given situation. 
When the agent learns a new skill, its initial expected value function is based on the single demonstration. 
On subsequent problem solving, the agent applies learned skills that have positive expected value and receives correctness feedback (coded as a reward of +1 or a penalty of -1). This feedback is used to refine the skill's expected value function. For example, the agent might learn that the new add-and-copy skill only produces a reward when applied to two numbers that have an addition sign between them. \citet{maclellan2017thesis} provides more details on the Apprentice Learner Architecture and its rationale.

\section{Tuning a Model to Account for Individual Student Differences}\label{sec:tuning}

In prior work \citep{maclellan2016apprentice,maclellan2017thesis}, we explored the use of Apprentice Learner models for predicting which fractions interventions yield the best learning. These studies successfully predict students will have lower error during tutoring in the blocked condition and lower error on a posttest in the interleaved condition. However, when we compared the learning curves generated by the humans and agents, we found a large discrepancy. Our previous work assumed that all students were identical learners that came to the fraction arithmetic task without any prior fraction arithmetic knowledge, so agents made an error on every first opportunity to apply a skill, as they always requested a hint on each first opportunity, which we counted as an error. In contrast, the data show that human students make mistakes on their first skill opportunities less than half the time, suggesting that students enter the tutoring setting knowing how to do most of the fraction steps.
Beyond differences in prior knowledge, we also anticipate that students have cognitive differences that impact their learning. For example, students might have varying thresholds for their willingness to guess an answer, which may affect how quickly they learn. Our previous approach did not account for these individual-level differences.

\subsection{Our Model Tuning Approach}

\begin{figure}[t]
\vskip 0.05in
\begin{center}
\includegraphics[width=\textwidth]{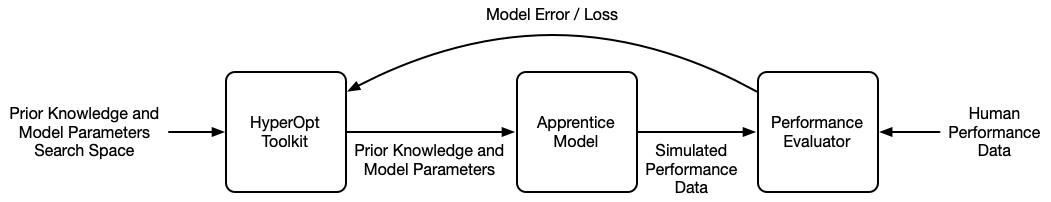}
\caption{The iterative process used to tune an Apprentice agents to better model a specific student.} 
\label{fig:hyperopt-process}
\end{center}
\vskip -0.2in
\end{figure} 

To address this gap, we devised an approach to automatically tailoring models of specific students to better account for their differences. We use the HyperOpt tookit \citep{hyperopt} to create personalized agents representing specific students, as shown in Figure \ref{fig:hyperopt-process}. Each agent is parameterized by its initial skills (prior knowledge), as well as cognitive parameters (e.g., its disposition towards exploration). 
We provide HyperOpt with the space of all skills and the range of values for each parameter. To tune to a specific student, HyperOpt iteratively samples skills and parameters. It creates an Apprentice Learner agent with these sampled values, simulates the target student and corresponding intervention, and evaluates how accurately the agent emulates the observed behavior on each step. The error is fed back to the HyperOpt toolkit, which uses Bayesian inference to update its sampling distribution for selecting skills and parameters. With each iteration, HyperOpt converges towards a configuration that minimizes the differences with the student.

Our tuning process bears some resemblance to knowledge tracing. 
For example, \citet{conati-student-modeling} explore the use of sampling to update a Bayesian model of each student's knowledge state.
Also, Corbett and Anderson (\citeyear{corbett1994knowledge}) used a Bayesian sequence model (a hidden Markov model) to estimate each student's knowledge at each time point.
One central difference from our approach is how they represent expertise.
Knowledge tracing has only an abstract representation of skills (e.g., the probability that a student knows a given skill and would uses it correctly).
In contrast, our approach explicitly represents each skill as a rule-like structure within the Apprentice Learner architecture (see Section \ref{sec:apprentice}).
Additionally, the other approaches maintain a best estimate of each student's knowledge state over the course of problem solving, whereas ours focuses on generating a best estimate of a student's initial knowledge state.
Finally, our approach uses a mechanistic model of learning, embodied in the Apprentice Learner agents, that the others lack.


To conduct tuning, we created prior knowledge to initialize the process. In our previous work, all agents had expertise in whole number arithmetic (adding, subtracting, multiplying, and dividing two numbers). The knowledge space included these skills and fraction arithmetic skills for adding fractions with the same denominators, converting fractions to common denominators with the butterfly method, and multiplying fractions. If an agent starts with all the fraction arithmetic skills, then it will get every step correct without practice. However, if it has only some of them, then it will get some steps correct and have to learn others through worked examples and practice. Our agents also had parameters for reinforcement learning, including how often to guess random actions, the penalty for taking an action (to minimize unnecessary actions), and the decay in reward when propagating it back over decisions.

\subsection{Evaluation of Model Tuning}

\begin{figure}[!t]
\vskip 0.05in
   \centering
   \includegraphics[width=\textwidth]{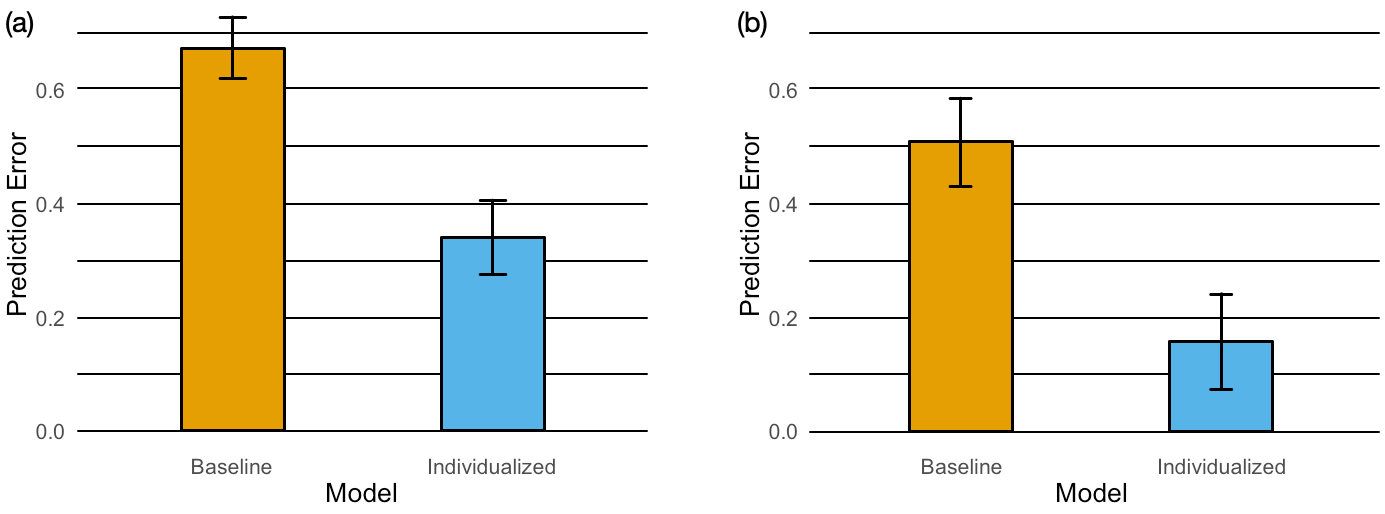}
   \caption{(a) Prediction Error for Apprentice Learner Models that are trained and tested on the same data. (b) Prediction Error for Apprentice Learner Models that are trained on earlier data and tested on later unseen data (within individuals). In both cases, the error bars represent the 95\% confidence intervals.}
   \label{fig:tuned-vs-untuned}
   \vskip -0.2in
\end{figure}

We conducted two evaluations of this approach. First, we examined whether HyperOpt could identify skills and parameters that improved the alignment between each agent and its target human. We tuned 24 Apprentice Learner agents to fit 24 students using 20 iterations of HyperOpt optimization in each case. For the performance evaluation step of the tuning process (Figure \ref{fig:hyperopt-process}), we simulated each student's behavior on their first ten tutor problems and analyzed how frequently the simulation correctly predicted the correctness of the respective student's first attempt on each step. 

As an example, imagine we applied our approach a high-performing student that already knows all fraction arithmetic skills.
Initially, HyperOpt might select an agent configuration that has expertise in whole number arithmetic but not fractions.
The system would construct an Apprentice agent using this configuration and apply it to simulate problem solving and learning on the first ten problems the human received from the tutor.
HyperOpt's sampling distribution would then be updated based on the similarity between the agent and human performance.
In this case, the agent would perform worse than the human, so HyperOpt's sampling distribution would be updated to favor different configurations.
Eventually, the system would converge to a configuration with prior knowledge of all fraction arithmetic, as this best replicates the human's behavior.

After fitting a model to each student, we compared the tuned agents to a baseline that had whole number arithmetic but not fraction arithmetic and default cognitive parameters (our default model guesses random actions 30\% of the time, receives a penalty of $0.05$ for every action, and decays future rewards by $70\%$ over each time step). This baseline corresponds to the configuration used in our previous work \citep{maclellan2016apprentice,maclellan2017thesis}. For each agent, we computed the error between it and the corresponding student on the first ten problems received in the tutor. Figure \ref{fig:tuned-vs-untuned}(a) shows the results of this study. The main finding is that the tuned models have lower error than the baseline and better approximate the human learning trajectories. 

This result demonstrates that our tuning approach reduces the error between agents and specific students. However, it evaluates each on the tutor problems that were used to tune it---training and testing on the same data. To assess how well the tuning improves predictions for the target students on unseen data, we used a form of temporal cross validation. We constructed tuned models for 15 students by using HyperOpt to reduce the error with the respective humans on the first five problems. We then compared these tuned agents to our baseline on the next 15 problems that they received from the tutor (problems 6 through 20), which were not used as part of the tuning process. Figure \ref{fig:tuned-vs-untuned}(b) shows the results of this comparison. We find that tuning also yields better predictions on unseen data that were not used for that purpose. It is worth noting that the prediction error in this second evaluation is lower than the first because students' errors tend towards zero as they receive more fractions practice, so predicting the human performance becomes easier at later opportunities.

\section{Generating Counterfactual Prediction with Personalized Models}

Although \citet{patel2016block} argue for a one-size-fits-all approach to problem ordering---that interleaved practice yields better learning than blocked practice---we take a more nuanced view that different kinds of practice are better for different students, depending on their prior knowledge and other differences. Unfortunately, this view complicates the intervention design problem because it means the designer must find the best intervention for each student, rather than a single intervention that works best for everyone. Having demonstrated that personalized models generate improved predictions for their target students, we explored how they might inform the selection of which tutoring interventions will be best for each recipient.

To support instructional design, our approach can generate counterfactual predictions of a student's performance. If previously collected intervention data for a student are available, then a designer can create a personalized Apprentice Learner agent with these data and apply it to predict what the student's performance would have been in the counterfactual condition. The model could also be tuned with other information, such as pretest results. By leveraging such data, the designer can generate counterfactual predictions about how a particular student would respond to an intervention prior to administering it.

\begin{figure}[t]
\vskip 0.05in
\begin{center}
\includegraphics[width=\textwidth]{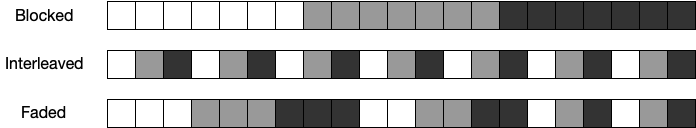}
\caption{Conceptual depiction of three possible orderings for the fraction arithmetic tutor. Shadings of cells denote the three types of problems: fraction addition with same denominators, fraction arithmetic with different denominators, and fraction multiplication.} 
\label{fig:sequences}
\end{center}
\vskip -0.2in
\end{figure} 

In this paper, we examined such counterfactual predictions. In particular, we explored how our Apprentice Learner agents might be applied to answer four counterfactual questions regarding students who used the fraction tutor:
\vskip 0.05in

\cbullet 
Q1: What learning curves would result if a participant in the interleaved condition had received the blocked condition and vice versa?

\cbullet 
Q2: What would have happened if a participant's problems were initially blocked but faded to interleaved over the course of instruction?

\cbullet 
Q3: What would a low-performing student's learning curves look like for each intervention?

\cbullet
Q4: What would a high-performing student's learning curves look like for each intervention?

\vskip 0.05in
\noindent
The first question concerns what would have happened if a student received a different condition from the one that was randomly assigned. The second question investigates how interventions in the human study would compare to a novel faded intervention for which no data are available. This would start out like the blocked instruction, but then fade into interleaved instruction over time. Figure \ref{fig:sequences} provides a conceptual depiction of the three instructional interventions. We explored this faded blocked to interleaved intervention because previous work \citep{carvalho2015benefits} shows that blocking and interleaving support different kinds of learning---blocking helps students learn which task features are relevant but interleaving helps them discriminate among competing skills. Finally, the third and fourth questions address how alternative interventions might differentially affect high and low-performing students. In general, we feel these counterfactual questions are representative of what an intervention designer might ask.



\subsection{Evaluating the Use of Personalized Models for Counterfactual Prediction}

To evaluate our approach, we tested its ability to answer the four counterfactual questions. 
For each query, we started by tuning models to target students using performance data from the interventions they received, as described in Section \ref{sec:tuning}. We then simulated each student's behavior in counterfactual interventions using the respective personalized agent to determine what each of them would have done had they received different interventions.

We focused our simulation efforts on modeling two specific students from the human data. We will refer to them as the `high-performing student' and the `low-performing student.' The first had the highest tutor performance of those in the blocked condition and the second had the lowest tutor performance of those in the interleaved condition. For each one, we constructed a personalized agent using their performance data. 
We chose to generate counterfactual predictions for these two students because we believed they would exhibit clear differences (e.g., in their prior knowledge), which would enable us to showcase the value of our personalization approach. 

After tuning agents to each student, we applied them to simulate three different counterfactual interventions. First, we used variations of the conditions administered, blocked or interleaved, where each variation had a randomized problem ordering within the target ordering schema. To see how well our model predicted the learning trajectory for the observed condition, we compared the simulated behavior on these different variations to the corresponding student's performance. Next, we simulated the performance on the opposite condition. Finally, we simulated behavior on the `faded blocked to interleaved' condition. For each counterfactual condition, we simulated behavior 20 times. For each iteration in the blocked and interleaved conditions, we randomly selected a problem ordering from the sequences that were actually administered to people in that condition from the larger data set. For the faded condition, we randomly generated sequences in which learners received problems in blocks of three (three addition with same denominators, then three addition with different denominators, then three multiplications), then blocks of two, and eventually blocks of one. Across these evaluations, our agents generated predictions for interventions that differed from those actually experienced by the humans, so no ground truth was available (i.e., they were counterfactual predictions). 

\subsubsection{High-Performing Student Predictions}


\begin{figure}[!t]
\vskip 0.05in
   \centering
   \subfloat[][]{\includegraphics[width=.33\textwidth]{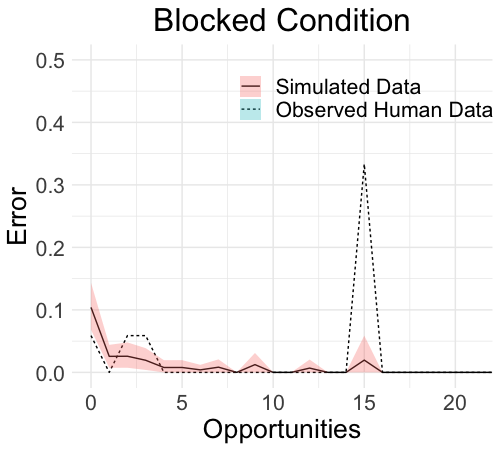}}
   \subfloat[][]{\includegraphics[width=.33\textwidth]{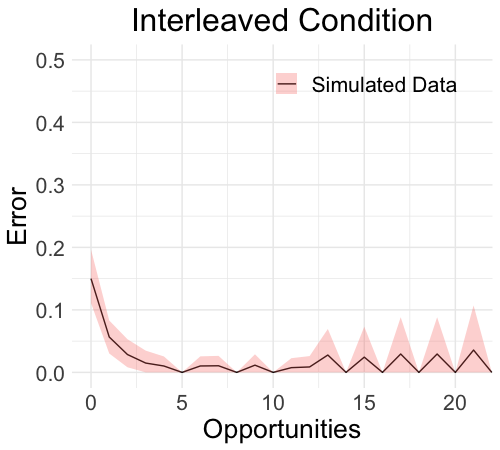}}
   \subfloat[][]{\includegraphics[width=.33\textwidth]{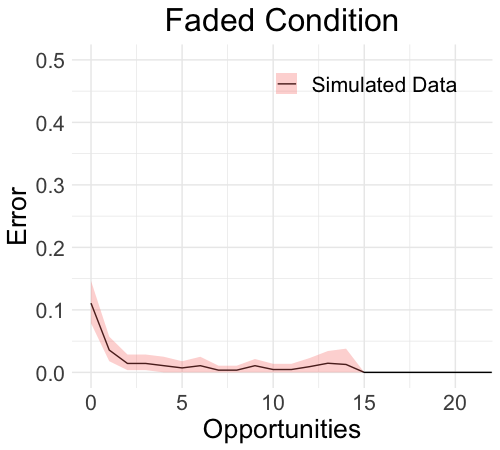}}
   \caption{The high-performing student's predicted and actual performance in the blocked condition (a), as well as their predicted performance in the interleaved (b) and the faded blocked to interleaved conditions (c). Shaded regions denote the 95\% confidence intervals for predicted error at each opportunity.}
   \label{fig:high-performing}
\end{figure}

Figure \ref{fig:high-performing} shows the observed and simulated learning curves for the high performer in each of the three interventions (blocked, interleaved, and faded), reporting average performance across all fraction arithmetic skills. The opportunity count represents how many prior opportunities the learner had to exercise each skill. Thus, the error at opportunity zero corresponds to the average performance the first time the student applied each skill within the tutor. For these plots, we treated each input field for each problem type as exercising a unique skill; for example, filling in the numerator for an addition problem with the same denominators was a different skill than specifying the numerator for a multiplication problem. 

If we compare the actual human performance to the predictions for how the student would perform in variations of that condition, as in Figure \ref{fig:high-performing}(a), we find our approach reasonably emulates the observations. In particular, the large discrepancy between error rates on the first opportunity, observed in prior Apprentice Learner work, is absent here. Previous models had 100\% error rate on the first opportunity because they always started without any fraction knowledge and were not tailored to specific students \citep{maclellan2016apprentice,maclellan2017thesis,Weitekamp2019zeroparam}.
The new results suggest we can overcome this disagreement by taking into account each student's unique knowledge and cognitive parameters.

Additionally, we see that the model correctly predicts a spike in error rate around opportunity 15, likely due to a transition between problem blocks at this point.\footnote{Figure \ref{fig:high-performing}(a) shows data from one human student on a single sequence, whereas the simulated data were computed over 20 students on 20 sequences; this is why there are no error bars and why the error rate on opportunity 15 appears to be much higher.} The predicted performance in the interleaved condition, in Figure \ref{fig:high-performing}(b), reveals an interesting spiky pattern in error in the tail of the learning curve as the student alternates among problems of different types. Our agent also predicts that error in the interleaved condition will be higher than in the blocked condition. This agrees with  Patel et al.'s (\citeyear{patel2016block}) finding that students make more errors in the interleaved condition than in the blocked condition. The predicted performance in the faded condition, shown in Figure \ref{fig:high-performing}(c), is almost identical to that in the blocked condition, even though the problems are essentially interleaved at the higher opportunity counts. Interestingly, the model does not predict spikes in error here as it did for the interleaved condition. This suggests that the fading strategy combines the benefits of blocking and interleaving, achieving better learning and perhaps better transfer to a posttest, although we did not test this idea.

\subsubsection{Low-Performing Student Predictions}


\begin{figure}[!t]
\vskip 0.05in
   \centering
   \subfloat[][]{\includegraphics[width=.33\textwidth]{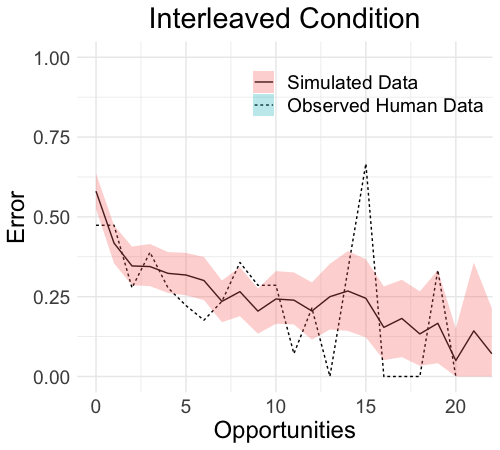}}
   \subfloat[][]{\includegraphics[width=.33\textwidth]{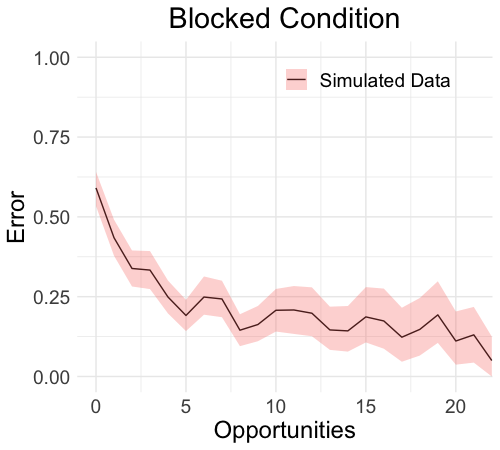}}
   \subfloat[][]{\includegraphics[width=.33\textwidth]{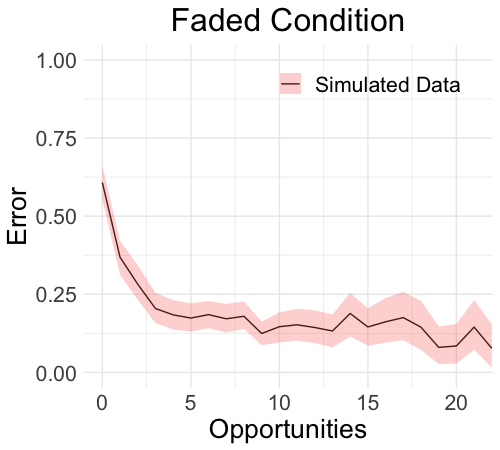}}
   \caption{The low-performing student's predicted and observed performance in the interleaved condition (a), as well as their predicted performance in the blocked (b) and the faded blocked to interleaved conditions (c). Shaded regions denote the 95\% confidence intervals for predicted error at each opportunity.}
   \label{fig:low-performing}
\end{figure}

As signified by a low error rate (less than 10\% when applying skills for the first time), the high performer starts training with ample prior fraction knowledge. To explore behavior at the other end of the spectrum, we constructed a personalized model of the low performer, who we hypothesize has more limited expertise.
Figure \ref{fig:low-performing} shows the learning curves (simulated and observed) for this student in each possible intervention. As expected, this learner makes many more errors than the other one, but still gets approximately half of their first steps correct (initial error rate around 48\%), suggesting a fair amount of prior fractions knowledge. 

Figure \ref{fig:low-performing}(a) compares the student's observed performance on the interleaved condition with the model's predictions on its variations. The graphs show that the model generates reasonable predictions that are in alignment with the observations. There is only a minor difference on the first opportunity, with the model predicting a slightly higher error. As mentioned previously, this discrepancy is not as large as that observed in previous Apprentice Learner work, where the model predicts 100\% error on the first step. Additionally, the observed performance is generally within the predicted confidence intervals generated by the model. If we compare the predicted interleaved performance in Figure \ref{fig:low-performing}(a) to the predicted blocked performance in Figure \ref{fig:low-performing}(b), we see that the model predicts a lower error rate in the latter than in the former, which agrees with predictions for the high-performing student and  general experimental findings \citep{patel2016block}. Finally, the predicted performance for the faded condition in Figure \ref{fig:low-performing}(c) has a lower overall error and a faster decrease than either the interleaved or blocked conditions. In particular, error in the faded condition decreased to approximately 20\% by opportunity 3, whereas this level is not achieved until opportunity 15 in the interleaved condition or until opportunity 5 in the blocked condition. We did not find the same improvement for the high-performing student in the faded condition, but there was less range for possible improvement. This finding further suggests that the faded condition combines the benefits of blocking and interleaving.

\subsection{Discussion}

These results demonstrate how intervention designers might use personalized models to predict how students would respond to counterfactual interventions. This capability is powerful because it should let designers conduct low-cost simulations to evaluate many alternative interventions. They can then run human studies to evaluate the interventions that the simulations suggest will be most promising. Before designers can trust predictions from Apprentice Learner models, they need evidence that the counterfactual predictions approximate human performance. Unfortunately, evaluating such predictions using previously collected performance data is difficult because, by definition, no ground truth data are available. However, future work can and should explore the design of experiments to test our predictions.

We claim that our findings provide evidence that our models make reasonable counterfactual predictions about how students will respond to alternative interventions. For both the high and low performers, they generated understandable predictions. When we compared the observed performance to the predictions for different training regiments, we found a close agreement. Further, for both students they predicted lower error rates in the blocked than interleaved condition, which has been observed in prior studies \citep{patel2016block}.

It is harder to evaluate the model predictions for the faded condition because no human data are available. However, our approach makes a reasonable, but not entirely obvious, prediction that students in the faded setting will have lower overall error and that it will decrease more quickly than either in the blocked or interleaved conditions. Additionally, it predicts that error in the tail of the learning curve will look more like that in the blocked condition (less spiky) than in the interleaved condition (more spiky), even though practice at higher opportunities in the faded condition is essentially interleaved. We have not yet evaluated these predictions with human experiments, but we argue they constitute reasonable counterfactual predictions that are consistent with prior research on blocking versus interleaving \citep{carvalho2015benefits}. A good test of our approach would be to run a human study comparing faded problem ordering to the blocked and interleaved conditions.

Finally, these simulated counterfactual data provide some answers to our four counterfactual questions. They give a picture of how students' learning curves might differ if they were in the blocked or interleaved condition (Q1) and they show how students' learning curves might differ if they were in a novel faded condition (Q2). They also provide a picture of how low and high performers would respond differently to these interventions (Q3 and Q4). In general, our study suggests that which intervention the high performer receives matters little due to their ample prior fractions knowledge. However, the low performer improves in all three conditions, but seems to improve the most in the faded condition.

\section{Related Research}

Our work is not the first to explore the application of computational models to guide the design of interactive systems. \citet{card1986model} proposed using a Model Human Processor that encapsulates cognitive theory into a computational approach to evaluate the usability of interface designs in lieu of more costly human experiments. More recently, \cite{john2004predictive} have worked to realize this vision through the development of the CogTool system, which supports designers in building usable interfaces.
One key limitation of these efforts, with respect to the current work, is their emphasis on expert rather than novice performance. Additionally, they have not been applied to learning environments, as they possess no models of the learning process.

Several lines of research aim to model students within tutors. A large portion has focused on knowledge tracing \citep{conati-student-modeling,corbett1994knowledge,maclellan2015accounting}, which only abstractly (not mechanistically) models the learning and decision-making process. As a result, it requires previously collected data to fit parameters before it can generate predictions---it cannot generalize to interventions for which no data are available.
In contrast, computational models of learning, such as Cascade \citep{vanlehn1992cascade}, STEPS \citep{ur1995steps}, SimStudent \citep{li2012efficient}, and Apprentice Learner \citep{maclellan2017thesis}, possess mechanistic models of learning and decision making---they model how knowledge is applied to generate behavior and how it is updated in response to examples and feedback.
As a result, they can generate purely theory-driven predictions, even for interventions that lack student data.
Our work builds directly on these projects to explore how they can support instructional designers in evaluating the effectiveness of different training interventions, similar to how \citet{card1986model} and \citet{john2004predictive} proposed using computational models to guide interface design.

Other researchers have also investigated how to personalize cognitive models to specific people. For example, Jones and VanLehn (\citeyear{jones1992fine}) explored how to hand tune the prior knowledge and parameters of Cascade models to align them with student protocols. More recently, Zhang and Hornof (\citeyear{zhang2014understanding}) used large-scale simulation of all possible variations in prior knowledge and parameter configurations to approximate (and explain) individuals' behaviors and strategies.
\citet{Weitekamp2019zeroparam} tuned Apprentice Learner agents to specific students using a more implicit approach. Rather than searching over the space of prior knowledge, they statistically estimated how much previous practice each student had with each problem type and pretrained Apprentice Learner agents on an equivalent number of comparable problems. Our method builds on this work to tune prior knowledge and parameters to target students. However, we chose to estimate knowledge directly rather than use the amount of previous practice. We also chose automated rather than manual tuning, and leveraged HyperOpt \citep{hyperopt} to guide search instead of simulating all possible configurations. More work is needed to compare our explicit model tuning method to Weitekamp et al's (\citeyear{Weitekamp2019zeroparam}) implicit approach. A nice feature of our system is that it generates interpretable prior knowledge, which the designer can inspect. However, it requires that we specify the set of knowledge to search over. In contrast, pretraining estimates how much prior practice students had with each problem type rather than explicitly representing their prior knowledge.

\section{Conclusions and Future Work}

This paper has presented evidence to support three high-level claims: (1) that computational models of learning can support causal prediction of human behavior in response to training interventions; (2) that one can tune these models to better predict performance for individuals by adjusting prior knowledge and parameters; and (3) that these personalized models generate plausible counterfactual predictions for target students. To support these claims, we extended the Apprentice Learner Architecture and demonstrated its use for causally reasoning about which fraction arithmetic interventions would produce better learning in particular students. We described how to tune agents to individuals using performance data and showed that they predict student performance better than generic agents. Finally, we constructed personalized models for high and low performers, then used them to counterfactually predict learning curves for three different interventions. The results showed that personalization yields reasonable predictions that agree with the available human data. The approach also generate plausible predictions for a novel intervention for which no data are available but that might be tested with future experiments.

There are many possible directions for future research. 
We are particularly interested in preregistering our predictions for the `faded blocked to interleaved' intervention and running a study with humans to test them. 
Specifically, our work suggests that the faded condition will yield faster learning and less spiky error rates in the learning curve's tail.
Future investigations should also administer quizzes to agents at the end of training to evaluate final performance across all skills.
\citet{patel2016block} showed that learners have lower error on a posttest in the interleaved condition, even though they did better during tutoring in the blocked condition.
This additional evaluation will let us investigate these effects using simulated agents.


In previous research \citep{maclellan2020domain}, we emphasized the application of computational learning models to support tutor development in psychology, chemistry, math, language learning, and engineering. 
However, most prior work has focused on mathematical tasks.
Future efforts should explore use of our approach on additional domains to showcase its generality.
Additionally, past research has focused almost exclusively on learning within intelligent tutoring systems and we need more investigations of how the approach applies to other settings, such as educational games.
To support these environments, our framework will likely require substantial extensions. For example, our agents currently only support step-based interactions and expect immediate feedback, but most games require continuous-time interaction and provide only delayed feedback.

We should also expand the framework to account for additional learning phenomena, such as the testing effect, where  students' performance improves after taking tests, even though they receive no feedback or instruction.
Modeling this effect would require extending agents to engage in learning even in the absence of such information.
We hypothesize that agents will need mechanisms to generate their own internal feedback when taking tests, which they can use to refine their skills during test taking.
Future work should explore variations of this idea to account for how students learn when engaging in problem solving without external feedback or guidance.

Finally, although there have been some efforts to make our approach accessible to training intervention developers, such as teachers or instructional designers \citep[e.g.,][]{weitekamp2020interaction}, we need more work in this area. 
Future research should collaborate with potential end users to design interfaces that are usable and do not require technical expertise to automate the steps currently being handled manually on our team.
We should explore approaches that support instructional designers by automatically searching the space of training interventions, say by searching the space of problem orderings to find ones that predict the best learning.

 
\begin{acknowledgements} 
\noindent
The first author started this work at Soar Technology and continued it at Drexel University and then Georgia Institute of Technology. The other authors began this research at the University of Alabama and continued it at their currently listed institutions. Our study was funded in part by Award HR00111990055 from the DARPA TAILOR program, Award 2112532 from NSF's AI-ALOE Institute, and Award W911NF2120101 from ARL's STRONG program.
The views, opinions, and/or findings expressed are
those of the author and should not be interpreted as representing
the official views or policies of the funding agencies.
We thank Rob Sheline, Ben Mastay, and Frank Koss their technical development support and Pat Langley for his feedback and comments on earlier drafts.
\end{acknowledgements} 


{\parindent -10pt\leftskip 10pt\noindent
\bibliographystyle{cogsysapa}
\bibliography{paper}

}


\end{document}